\documentclass{article}

\usepackage[preprint]{neurips_2026}

\usepackage[utf8]{inputenc} 
\usepackage[T1]{fontenc}    
\usepackage{hyperref}       
\usepackage{url}            
\usepackage{booktabs}       
\usepackage{amsfonts}       
\usepackage{nicefrac}       
\usepackage{microtype}      
\usepackage{xcolor}         

\usepackage{tikz}
\usetikzlibrary{arrows.meta, decorations.pathreplacing, positioning, calc}
\usepackage{amsmath}
\usepackage{multirow}

\title{Logit Averaging During Training Sets a New State of the Art for Non-Invasive Speech BCI}

\author{%
  Miran Özdogan \\
  University of Oxford\\
  \texttt{miran@robots.ox.ac.uk} \\
  \And
  Gereon Elvers \\
  University of Oxford\\
  \texttt{gereon.elvers@tum.de} \\
  \And
    Michael Bronstein \\
    University of Oxford\\
    \texttt{michael.bronstein@cs.ox.ac.uk} \\
  \And
  Oiwi Parker Jones \\
  University of Oxford\\
  \texttt{oiwi@robots.ox.ac.uk} \\
}

\begin{document}

\maketitle

\begin{abstract}
Non-invasive brain-computer interfaces are in practice limited by signal-to-noise ratio, and recent state-of-the-art methods address this limitation by aggregating over repeated trials of the same stimulus. Yet multi-trial aggregation has been treated inconsistently across the literature: some work averages brain signals at the model's input, other work averages predictions at the model's output, and neither treats aggregation as an architectural choice in its own right. We provide the first systematic comparison of aggregation strategies for non-invasive neural decoding, evaluating six paradigms (including logit averaging during training and three attention-based variants we introduce) across three datasets, four tasks, and two group sizes. We find that aggregation choice matters substantially: training-time logit averaging, the simplest method we propose, outperforms all others in 6 of 7 settings. On the phoneme classification setting of the 2025 PNPL competition, it sets a new state of the art, significantly outperforming the competition winner by 17.4 absolute percentage points in balanced accuracy. We further draw a theoretical connection between logit averaging and joint ensemble training, which has been shown to underperform independent ensembling in the deep learning literature; our results call into question recent explanations of why joint ensemble training fails. Together, these findings establish multi-trial aggregation as a key design choice for non-invasive neural decoding and identify training-time logit averaging as a simple, effective default.
\end{abstract}

\section{Introduction}

Driven by advances in deep learning, brain-computer interfaces (BCIs) have made substantial progress in recent years. Speech BCIs in particular have begun to restore communication in people with anarthria, decoding attempted speech~\citet{moses2021nejm, makin2020, metzger2023neuroprosthesis, willett2023high, card2024nejm}, imagined or inner speech~\citet{Kunz2025inner}, and attempted handwriting~\citet{Willett2021handwriting} from cortical activity. The most recent intracranial systems have reached word error rates below 5\%, while approaching the speed of natural conversation ~\citet{card2024nejm}.

These achievements have, however, relied on invasive recordings that require neurosurgical implantation. Both the surgical procedure itself and its associated long-term risks restrict the deployment of these systems. Motivated by this constraint, a growing body of work has investigated whether comparable performance can be achieved from non-invasive measurements~\citet{jayalath2024scaling, jayalath2025cracking, dascoli2024, defossez2023, landau2025competition, ozdogan2025libribrain}.

Among non-invasive modalities, MEG-based approaches have produced the strongest recent results~\citet{dascoli2024}. However, MEG signals carry a low signal-to-noise ratio relative to intracranial recordings, and single-trial decoding accuracies remain far below the levels required for practical communication. Recent work has shown that this gap can be partially bridged by aggregating predictions across multiple trials of the same stimulus~\citet{Petit2020,ozdogan2025libribrain,landau2025competition, dascoli2024}: averaging across trials reduces noise and pushes balanced accuracies on tasks such as phoneme classification well above what is achievable from single trials.
\begin{figure}[t]
\centering
\includegraphics[width=\linewidth]{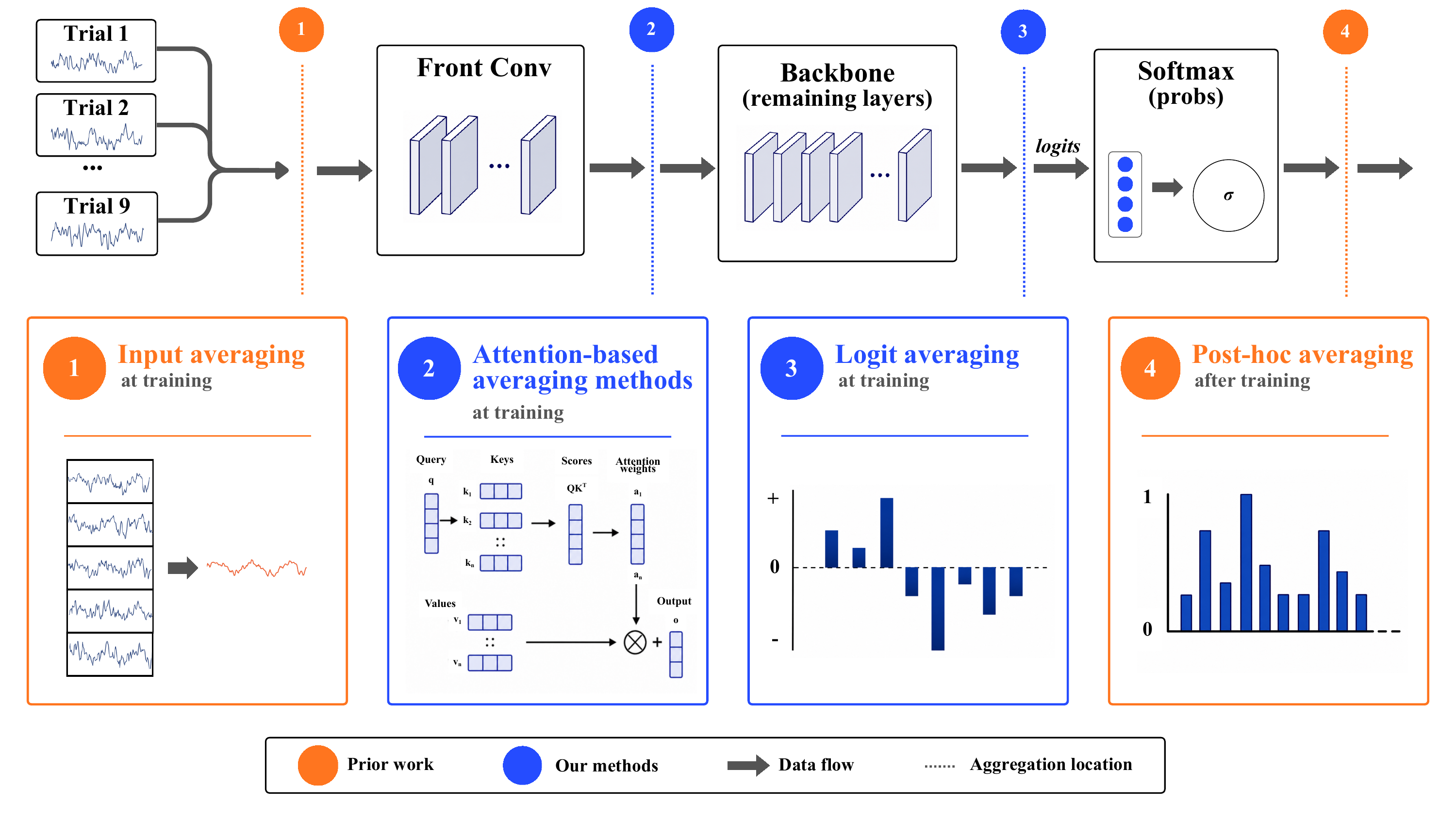}
\caption{\textbf{Multi-sample aggregation paradigms.} A group of $g$ trials elicited by the same stimulus passes through a shared CNN backbone. The paradigms compared in this paper differ in \emph{where} along the pipeline aggregation across trials occurs: at the input (prior work), at intermediate features via learned attention or weighting (ours), at the logits during training (ours), or after softmax at evaluation time (prior work).}
\label{fig:paradigms}
\end{figure}

Yet multi-sample aggregation has been treated inconsistently across the literature: some work averages brain signals at the model's input~\citet{Petit2020, ozdogan2025libribrain, landau2025competition}, other work averages predictions at the model's output~\citet{dascoli2024}, and neither treats aggregation as a design decision in its own right. Both approaches also treat samples as independent, averaging either raw signals or independent predictions. Intuitively, more adaptive aggregation methods that let the model exploit structure across samples should perform better~(\ref{fig:paradigms}). 

We provide the first systematic study of aggregation as a design choice, comparing six paradigms (including four we propose) across three datasets and four tasks. We find that aggregation choice matters substantially: training-time logit averaging, the simplest method we propose, outperforms all others in 6 of 7 settings. On the phoneme classification setting of the 2025 PNPL competition, it sets a new state of the art, significantly outperforming the competition winner by 17.4 absolute percentage points in balanced accuracy. We further connect logit averaging to recent theoretical analyses of joint ensemble training and argue that our results call into question recent explanations of why joint training fails.

Our contributions are as follows:
\begin{itemize}
    \item We propose training-time logit averaging, a small change to the training objective, as a method for multi-trial aggregation. On the 2025 PNPL competition phoneme classification benchmark, it sets a new state of the art, significantly outperforming the competition winner by 17.4 absolute percentage points in balanced accuracy.
    \item We provide the first systematic comparison of aggregation strategies for non-invasive neural decoding, evaluating six paradigms across three datasets, four tasks, and two group sizes. Logit averaging emerges as the clear winner, outperforming other methods in 6 of 7 settings.
    \item We connect logit averaging to recent theoretical analyses of joint ensemble training and argue that our results call into question recent explanations of why joint ensemble training fails.
\end{itemize}

\section{Related work}
\subsection{Non-invasive speech decoding}

Recent years have seen rapid progress in decoding speech from non-invasive brain recordings. \citet{ozdogan2025libribrain} released the LibriBrain dataset, comprising over 50 hours of MEG from a single participant listening to naturalistic English speech, approximately five times larger than the next-largest within-subject MEG corpus by \citet{Armeni2022}. While many prior MEG speech-decoding datasets focus on multi-subject recordings of shorter duration per participant~\citet{gwilliams2022megmaschighqualitymagnetoencephalographydataset}, LibriBrain enables single-subject decoding at scale. This dataset underpins the 2025 PNPL competition~\citet{landau2025competition}, whose phoneme classification track evaluates submissions on samples formed by averaging 100 trials of the same phoneme.

\citet{dascoli2024} introduced a deep learning pipeline for decoding individual words from MEG and EEG, evaluated across 723 participants. Their analysis identified MEG as easier to decode than EEG, and observed that decoding performance scales with the amount of data per participant and with the number of trials averaged at test time. In contrast to the \emph{input averaging} used in the LibriBrain phoneme task, they apply \emph{post-hoc averaging} at evaluation time, averaging the predicted probability distributions across trials of the same word.

Building on these resources, \citet{jayalath2024scaling} scaled self-supervised pre-training and reported improvements over prior models on speech-decoding benchmarks. Most recently, \citet{jayalath2025cracking} extended this line of work to full sentence-level brain-to-text by combining word-classification models with LLM-based rescoring. Even with these methods, the best reported word error rate remains at 88\%, far from the levels required for practical deployment. Similar gaps exist for other non-invasive modalities, including fMRI~\citet{Tang2023} and EEG~\citet{yang2024neuspeechdecodeneuralsignal}. Together, these results underscore that single-trial decoding from non-invasive recordings remains bottlenecked by signal-to-noise ratio and that multi-trial aggregation is essential for achieving usable performance.

\subsection{Multi-sample prediction}

A multi-trial predictor takes as input a set of $g$ trials $\{x_1, \ldots, x_g\}$ elicited by the same stimulus and produces a single prediction $\hat{y}$ over the label space. Since the trials are interchangeable, we want our predictor $f$ to be \emph{permutation-invariant}: for any permutation $\pi$ of the indices $\{1, \ldots, g\}$, $f(\{x_{\pi(1)}, \ldots, x_{\pi(g)}\}) = f(\{x_1, \ldots, x_g\})$. This requirement has been studied extensively in the broader machine learning literature on set-structured inputs \citet{DIETTERICH199731, maron1997framework}. 

\citet{settransformer} introduced the Set Transformer, which aggregates set-structured data via attention while remaining permutation-invariant. Their work inspired the attention-based methods we evaluate in Section~\ref{sec:methods:proposed}, including a learned query attention mechanism in which a small fixed number of trainable query vectors attend over the set elements.

\subsection{Ensembling}
\label{sec:relatedwork:ensembling}

Multi-trial prediction is conceptually related to ensembling, a long-established technique in deep learning for reducing the variance of model predictions \citet{ hinton2015distillingknowledgeneuralnetwork, dietterichensemble, gal2016dropoutbayesianapproximationrepresenting, NIPS1994_b8c37e33, NIPS1997_59f51fd6}. The most common approach is the deep ensemble~\citet{lakshminarayanan2017simplescalablepredictiveuncertainty}, in which multiple copies of the same architecture are trained on identical data from different random initialisations, and their predictions averaged at test time. The theoretical motivation is to reduce the variance contributed to predictions by initialisation, since each ensemble member converges to a different minimum of the loss landscape.

Our setting is related but distinct. We train a single neural network rather than an ensemble: the same weights process every trial. However, we may think of each trial as the true underlying signal corrupted by per-trial noise sources, including unrelated brain processes and measurement noise.
Under this framing, passing $g$ trials through a single shared model under the post-hoc averaging approach (position 4 in Figure~\ref{fig:paradigms}) is structurally analogous to passing the true signal through an ensemble of $g$ base learners each consisting of a noising operation followed by the shared backbone. The variance being reduced under averaging is therefore variance in the input rather than variance from initialisation, but the basic statistical principle is the same.

A structural analogue to training-time logit averaging also exists in the ensembling literature. \citet{jeffares2023jointtrainingdeepensembles} examined whether ensembles can be trained jointly under a single loss applied to averaged predictions or averaged logits, rather than independently and aggregated at test time, and report that such joint training consistently underperforms independent ensembling. We discuss their theoretical considerations, and how they relate to our results, in Section~\ref{sec:discussion:ensembles}.

\section{Methods}
\label{sec:methods}

\subsection{Problem setup}

We consider the problem of predicting a discrete label $y \in \{1, \ldots, K\}$ (e.g.\ a phoneme or a word) from a group of $g$ corresponding neural recordings. Each recording, or \emph{trial}, is a multivariate time series $x \in \mathbb{R}^{C \times T}$ with $C$ sensor channels and $T$ time samples. A multi-trial predictor takes as input a group $G = \{x_1, \ldots, x_g\}$ and outputs a single prediction $\hat{y} \in \{1, \ldots, K\}$.

All paradigms we consider share a common backbone $f_\theta$ with parameters $\theta$, which can map a trial to per-trial logits $\ell_i = f_\theta(x_i) \in \mathbb{R}^K$. Per-trial probabilities are obtained as $p_i = \mathrm{softmax}(\ell_i)$. 

The six paradigms we compare in this paper differ in \emph{where} along the path from $G$ to $\hat{y}$ aggregation across trials occurs (\ref{fig:paradigms}). All paradigms are trained with cross-entropy loss, weighted by the inverse square root of class frequency to handle class imbalance. We evaluate all paradigms using balanced accuracy (the macro-average of per-class recall), reported on held-out test data.

\subsection{Baseline aggregation paradigms}
\label{sec:methods:baselines}

We compare against two aggregation paradigms drawn from prior work, illustrated as positions 1 and 4 in Figure~\ref{fig:paradigms}.

\paragraph{Input averaging.} The trials in a group are averaged element-wise before being passed through the backbone, producing a single set of logits per group:
\begin{equation}
    \bar{x} = \frac{1}{g} \sum_{i=1}^{g} x_i, \qquad \hat{\ell} = f_\theta(\bar{x}).
\end{equation}
The model is trained on averaged data. At evaluation time, the same procedure is applied: trials are averaged and a single forward pass produces the prediction $\hat{y} = \arg\max_k \hat{\ell}_k$. This is the aggregation strategy used in the LibriBrain phoneme classification task~\citet{ozdogan2025libribrain, landau2025competition}.

\paragraph{Post-hoc probability averaging.} Each trial is passed through the backbone independently, producing per-trial probabilities $p_i = \mathrm{softmax}(f_\theta(x_i))$. The model is trained on individual trials with standard cross-entropy loss; no grouping is used during training. At evaluation time, trials in a group are aggregated by averaging their predicted probabilities:
\begin{equation}
    \hat{p} = \frac{1}{g} \sum_{i=1}^{g} p_i, \qquad \hat{y} = \arg\max_k \hat{p}_k.
\end{equation}
This is the aggregation strategy used by~\citet{dascoli2024}.

\subsection{Proposed aggregation paradigms}
\label{sec:methods:proposed}

We introduce four aggregation paradigms that intervene at intermediate stages of the pipeline. The first aggregates at the logits during training; the remaining three aggregate at intermediate features within the backbone.

\paragraph{Logit averaging during training.} Each trial in a group is passed through the backbone independently, producing per-trial logits $\ell_i = f_\theta(x_i)$. The logits are then averaged across trials before the loss is computed:
\begin{equation}
    \hat{\ell} = \frac{1}{g} \sum_{i=1}^{g} \ell_i, \qquad \mathcal{L} = \mathrm{CE}(\hat{\ell}, y).
\end{equation}
Gradients flow through the averaging operation back to each per-trial forward pass, so the backbone is optimised to produce features whose logits are useful \emph{after averaging across trials}, rather than per-trial. At evaluation time, the same procedure is applied. This paradigm differs from post-hoc probability averaging only in that aggregation occurs at training time and over logits rather than over probabilities at evaluation time.

\paragraph{Attention-based aggregation methods.}
We additionally introduce three attention-based aggregation methods that aggregate per-trial features at the output of the backbone's first convolutional layer. \emph{Learned weighted averaging} computes per-trial scalar weights from each trial's features in isolation and combines the per-trial feature maps as a weighted average. \emph{Cross-trial attention} extends this with multi-head self-attention across trials, so that each trial's weight depends on all others. \emph{Learned query attention} uses trainable query vectors that attend over the per-trial features, with a residual connection to the elementwise mean. All three preserve permutation invariance by construction. Architectural details are given in Appendix~\ref{app:attention_methods}.

\section{Experiments}
\label{sec:experiments}

\subsection{Datasets and tasks}
\label{sec:experiments:datasets}

We evaluate the six aggregation paradigms on seven (dataset, task, group size) conditions spanning three datasets. A summary of all conditions is given in Appendix~\ref{app:datasets-table}. 

\paragraph{LibriBrain.} LibriBrain ~\citet{ozdogan2025libribrain} is a single-subject MEG dataset comprising over 50 hours of recordings while one participant listened to naturalistic audiobook recordings of Sherlock Holmes stories. Each recording is downsampled to 250\,Hz across 306 sensor channels. We evaluate three classification tasks derived from these recordings: \emph{Phoneme Classification} (predicting which of the 39 ARPAbet phonemes corresponds to a 0.5\,s    window aligned to phoneme onset), \emph{Phoneme Voicing} (a binary task collapsing the 39 phoneme classes into voiced versus unvoiced, on the same windows), and \emph{Word Classification} (predicting which of the ten most frequent words in the corpus corresponds to a 1.0\,s window aligned to word onset). The phoneme task is the setting of the 2025 PNPL competition, which evaluates submissions on samples formed by averaging 100 trials of the same phoneme.

\paragraph{\citet{Armeni2022}.} The Armeni et al. dataset comprises approximately 10 hours of MEG recordings per participant from three participants who listened to an English audiobook of Sherlock Holmes stories. Recordings use 269 sensor channels also downsampled to 250\,Hz. We evaluate two classification tasks derived from these recordings, matched to the LibriBrain conditions: \emph{Phoneme Classification} and \emph{Phoneme Voicing}. 

\paragraph{MegNIST.} MegNIST is a 10-class digit classification benchmark recorded with 306-channel MEG from a single participant who imagined saying a digit upon visual cue. Unlike the LibriBrain tasks, MegNIST involves imagined rather than heard speech, providing a complementary evaluation setting. We use precomputed splits of 10{,}000 training, 1{,}000 validation, and 1{,}000 test samples. The dataset is not yet publicly available but will be released in the future.

\paragraph{Group sizes.} For each task, we evaluate at group size $g = 10$. For LibriBrain phoneme classification we additionally evaluate at $g = 100$ to match the 2025 PNPL competition setting. We do not report $g = 100$ for the remaining tasks because the test sets contain too few samples per class to form reliable group-level evaluations at that size.

\subsection{Experimental setup}
\label{sec:experiments:setup}

\paragraph{Training protocol.} All paradigms are trained with the Adam optimiser without weight decay, using a class-weighted cross-entropy loss as described in Section~\ref{sec:methods}. Training runs for up to 200 epochs with early stopping on validation balanced accuracy and a patience of 10 epochs. We use the convolutional neural network architecture introduced by \citet{ozdogan2025libribrain}. For each combination of dataset, paradigm, and group size, we tune the learning rate by validation balanced accuracy. Details of the tuning procedure are given in Appendix~\ref{app:lr_tuning}.

\paragraph{Final evaluation.} For each (dataset, paradigm, group size) configuration we run the chosen learning rate with five random seeds. Statistical comparisons against the two prior-work baselines (input averaging and post-hoc averaging) are performed via two-sided paired $t$-tests on test balanced accuracy, paired by seed. For the phoneme classification setting at $g=100$, we additionally compare training-time logit averaging against the conformer architecture that won the phoneme track of the 2025 PNPL competition~\citet{landau2025competition}. Both methods are trained on the same data with matched random seeds and compared via a two-sided paired $t$-test on test balanced accuracy.

\subsection{Main results}
\label{sec:experiments:main}

\begin{table}[t]
\centering
\caption{Test balanced accuracy across three primary aggregation paradigms and seven (dataset, task, group size) conditions. Values are mean $\pm$ standard deviation across 5 random seeds. The best mean in each row is shown in \textbf{bold}. Significance markers indicate paired $t$-tests against input averaging: $^{\ast}$ for $p < 0.05$, $^{\ast\ast}$ for $p < 0.01$. Results for the three additional aggregation methods we evaluate are reported in Appendix~\ref{app:attention_results}.}
\label{tab:main_results}
\small
\setlength{\tabcolsep}{6pt}
\begin{tabular}{llccc}
\toprule
& & \multicolumn{2}{c}{\textit{Prior work}} & \textit{Ours} \\
\cmidrule(lr){3-4} \cmidrule(lr){5-5}
\textbf{Dataset} & \textbf{Task} & Input avg.\ & Post-hoc avg.\ & Logit avg.\ \\
\midrule
\multirow{4}{*}{LibriBrain}
& Phoneme ($g=10$) & $.176 \pm .008$ & $.076 \pm .003$ & $\mathbf{.241 \pm .011}^{\ast\ast}$ \\
& Phoneme ($g=100$) & $.565 \pm .028$ & $.113 \pm .014$ & $\mathbf{.641 \pm .051}^{\ast\ast}$ \\
& Voicing ($g=10$) & $.780 \pm .007$ & $.508 \pm .005$ & $\mathbf{.794 \pm .011}$ \\
& Word top-10 ($g=10$) & $.544 \pm .034$ & $.400 \pm .016$ & $\mathbf{.564 \pm .019}$ \\
\midrule
MegNIST & Digit ($g=10$) & $.432 \pm .048$ & $\mathbf{.526 \pm .043}$ & $.418 \pm .008$ \\
\midrule
\multirow{2}{*}{Armeni et al.}
& Phoneme ($g=10$) & $.213 \pm .004$ & $.079 \pm .002$ & $\mathbf{.240 \pm .006}$ \\
& Voicing ($g=10$) & $.790 \pm .005$ & $.524 \pm .005$ & $\mathbf{.791 \pm .004}$ \\
\bottomrule
\end{tabular}
\end{table}

Table~\ref{tab:main_results} summarises test balanced accuracy across the three primary aggregation paradigms. Two patterns emerge.

First, aggregation choice has a substantial effect on decoding performance. Differences between the best and worst paradigm exceed 50 absolute percentage points on phoneme classification at $g=100$ ($.641$ vs.\ $.113$) and remain large across all other conditions.

Second, training-time logit averaging is the dominant method, performing best on all conditions across LibriBrain and Armeni, with statistically significant gains over input averaging on phoneme classification at both group sizes and datasets. On MegNIST the ranking inverts: post-hoc averaging is the strongest paradigm and logit averaging among the weakest. The shift between the heard-speech and the imagined-speech task suggests that the optimal aggregation method depends on the structure of the task. We discuss possible explanations in Section~\ref{sec:discussion}.

The three attention-based aggregation methods we introduce do not consistently outperform logit averaging in our experiments. Full results are reported in Appendix~\ref{app:attention_results}; we return to this finding in the discussion.

\subsection{Comparison to the 2025 PNPL competition winner}
\label{sec:experiments:pnpl}

\begin{table}[t]
\centering
\caption{Comparison against the conformer architecture that won the phoneme track of the 2025 PNPL competition~\citet{landau2025competition} on the LibriBrain phoneme classification task at $g = 100$. Both methods are trained on the same data with matched random seeds. Values are mean $\pm$ standard deviation across 5 seeds. F1 macro is the metric used for the competition leaderboard. $^{\ast}$ indicates statistical significance vs.\ the conformer baseline at $p < 0.05$ (two-sided paired $t$-test).}
\label{tab:pnpl_comparison}
\small
\begin{tabular}{lcc}
\toprule
\textbf{Method} & \textbf{Balanced accuracy} & \textbf{F1 macro} \\
\midrule
Conformer (PNPL winner) & $0.467 \pm 0.048$ & $0.440 \pm 0.047$ \\
Logit averaging (ours)  & $\mathbf{0.641 \pm 0.051}^{\ast}$ & $\mathbf{0.605 \pm 0.058}^{\ast}$ \\
\bottomrule
\end{tabular}
\end{table}
Logit averaging during training improves over the conformer that won the phoneme track of the 2025 PNPL competition by 17.4 absolute percentage points in mean test balanced accuracy ($0.641 \pm 0.051$ vs.\ $0.467 \pm 0.048$, $p = 0.022$, $n = 5$).\footnote{We evaluate on the public LibriBrain test split, since the ground-truth labels for the competition holdout set are not publicly available. On this split, the winning conformer also underperforms input averaging applied to our backbone ($0.565 \pm 0.028$, see Table~\ref{tab:main_results}). This may indicate some degree of overfitting to the competition holdout via repeated feedback from the public leaderboard.} The improvement on F1 macro, the metric used for the competition leaderboard, is similarly substantial: 16.5 percentage points ($p = 0.026$, $n=5$). This establishes a new state of the art on the phoneme classification task at the group size used in the 2025 PNPL competition.

\section{Discussion}
\label{sec:discussion}

\subsection{Why does training-time logit averaging work?}
\label{sec:discussion:mechanism}

The improvement of 17.4 absolute percentage points over the 2025 PNPL competition winner does not arise from architectural innovation but from a change to the training objective: averaging logits across grouped trials before applying cross-entropy, rather than applying cross-entropy per trial. In the following, we discuss the mechanism by which this modified objective leads to improved generalisation.

Under standard per-trial cross-entropy, the loss for any individual trial saturates quickly as $\log(p_\text{true})$ approaches $0$. At $p_{\text{true}} = 0.6$, the gradient magnitude is half of its value at $p_{\text{true}} = 0.3$, and only one-sixth of its value at $p_{\text{true}} = 0.1$.

Under the logit-averaged cross-entropy objective, by contrast, a single trial continues to receive gradient signal until the \emph{averaged logits} assign high probability to the true class. Even if a single trial is already correctly predicted with high confidence, it continues to receive gradient if other trials in its group are not.

This is reflected in the per-trial logit distributions of trained models (Table~\ref{tab:logit_magnitudes}): under logit averaging, per-trial logits have substantially larger magnitudes, inducing sharper, lower-entropy probability distributions. The maximum softmax probability reaches $0.91$ on MegNIST and $0.66$ on phoneme classification, compared to $0.40$ and $0.08$ under post-hoc averaging. The margin between the maximum logit and the mean logit is $5$--$11\times$ larger for logit-averaged models across four tasks where we computed logit statistics.

The logit-averaging objective differs from per-trial cross-entropy in that the model is incentivised not to classify each trial individually, but to produce per-trial logits whose group-averaged prediction is correct. This appears to induce the model to learn per-trial representations that complement each other during averaged prediction. 

Intuitively, we might expect such complementary behaviour to also yield well-calibrated single-trial predictions: if a model accurately calibrates its confidence on each trial, low-confidence predictions can be more easily overruled by high-confidence predictions that are more likely to be correct. However, our calibration analysis (Appendix~\ref{app:calibration}) shows that this is not the case. While group-level predictions of logit-averaged models tend to be better calibrated than those of models trained with standard cross-entropy, single-trial calibration is consistently worse.

Single-trial balanced accuracy under logit averaging is nonetheless higher than under standard per-trial training on the phoneme and phoneme voicing task, suggesting that some of the performance gains of logit averaging stem from an improvement in training dynamics rather than from group-level coordination alone. Conversely, logit averaging exhibits worse single-trial accuracy than per-trial training on the word classification task, even though its group-level accuracy is higher, implying that group-level coordination also plays a meaningful role.

In summary, we conjecture that several mechanisms contribute to the success of logit averaging. The joint training objective appears to induce favourable learning dynamics that can improve even single-trial performance, while at the same time producing single-trial logits that tend to be better suited to group-level averaging. We believe these mechanisms warrant further investigation in future work.

\begin{table}[t]
\centering
\caption{Per-trial logit and probability statistics of trained models, evaluated on ungrouped trials. \emph{Max-mean margin} denotes the difference between the maximum logit and the mean logit per trial. \emph{Max softmax} denotes the probability assigned to the top-predicted class. Values are mean $\pm$ standard deviation across 5 seeds.}
\label{tab:logit_magnitudes}
\small
\begin{tabular}{lcccc}
\toprule
& \multicolumn{2}{c}{\textbf{Max-mean margin}} & \multicolumn{2}{c}{\textbf{Max softmax}} \\
\cmidrule(lr){2-3} \cmidrule(lr){4-5}
\textbf{Dataset} & Logit avg.\ & Post-hoc & Logit avg.\ & Post-hoc \\
\midrule
MegNIST          & $21.4 \pm 1.2$ & $2.14 \pm 0.27$ & $0.911 \pm 0.001$ & $0.401 \pm 0.035$ \\
Phoneme          & $11.9 \pm 0.9$ & $1.67 \pm 0.15$ & $0.663 \pm 0.026$ & $0.081 \pm 0.002$ \\
Phoneme voicing  & $2.09 \pm 0.18$ & $0.41 \pm 0.03$ & $0.900 \pm 0.007$ & $0.679 \pm 0.010$ \\
Word top-10      & $18.6 \pm 7.1$ & $1.60 \pm 0.19$ & $0.868 \pm 0.043$ & $0.317 \pm 0.026$ \\
\bottomrule
\end{tabular}
\end{table}

\subsection{Why do attention-based methods not consistently outperform logit averaging?}
\label{sec:discussion:attention}

The four methods we introduced span a range of expressivity. Logit averaging is the simplest: it imposes no parametric coupling across trials. Two multi-head attention-based methods, cross-trial attention and learned query attention, introduce learned interactions between per-trial features, with the goal of producing aggregated representations that exploit structure across trials. Despite this additional capacity, neither consistently outperforms logit averaging (Table~\ref{tab:main_results}).

Two interpretations are consistent with our results. First, the trials in our setting are by construction noisy realisations of the same stimulus, so the joint distribution of per-trial features is approximately exchangeable: every trial is informative in roughly the same way, and pairwise interactions between specific trials carry little information that uniform averaging would not capture. Modelling these interactions explicitly may therefore add capacity without adding signal, while introducing optimisation difficulties. Second, the attention modules are trained from scratch on dataset sizes ranging from $10^4$ to $10^6$ trials, modest by the standards of attention architectures \citet{vaswani2017transformer}, particularly given that all parameters operate over groups of $g \le 100$ items. Larger datasets may eventually reveal cases in which expressive aggregation is beneficial. 
We also note that the attention mechanisms we evaluated were not explicitly designed to encode an inductive bias toward sharing particular features that we believe may be useful for interpreting other trials.
Within the regimes we evaluate, the simpler bias of logit averaging is the more reliable choice.

\subsection{Relation to joint ensemble training}
\label{sec:discussion:ensembles}

Training-time logit averaging is structurally analogous to joint ensemble training discussed in Section~\ref{sec:relatedwork:ensembling}, in which a single loss is computed on the averaged predictions of multiple ensemble members rather than on each member individually. \citet{jeffares2023jointtrainingdeepensembles} report that this procedure consistently fails for deep ensembles, and explain the failure through the following decomposition, originally derived for regression by \citet{NIPS1994_b8c37e33}:
\begin{equation}
\mathrm{ERR}_\text{joint} = {\mathrm{ERR}}_\text{ind} - \mathrm{DIV}.
\label{eq:diversity_decomposition}
\end{equation}
The joint loss equals the mean loss of the individual members minus a \emph{diversity} term that captures how much each member's prediction deviates from the ensemble mean. \citet{NIPS1997_59f51fd6} extend this decomposition to classification ensembles trained by averaging pre-softmax scores, the setting structurally analogous to logit averaging, where the diversity term takes the form
\begin{equation}
\mathrm{DIV} = \sum_{j=1}^{M} \, D_{\mathrm{KL}}(y \,\|\, p_j) - D_{\mathrm{KL}}(y \,\|\, \bar{q}) = \sum_{j=1}^{M} \, D_{\mathrm{KL}}(\bar{q} \,\|\, p_j),
\label{eq:diversity_kl}
\end{equation}
with $y$ the target distribution, $p_j$ the predictive distribution of member $j$, and $\bar{q}$ the predictive distribution of the ensemble. 

\citet{jeffares2023jointtrainingdeepensembles} argue that joint optimisation enables base learners to inflate this diversity term during training in a way that does not generalise to held-out data, calling the phenomenon \emph{learner collusion}: base learners supposedly can coordinate to bias their predictive distributions in opposing directions that cancel out in aggregate, artificially maximising diversity across the training data.
However, from Equation~\eqref{eq:diversity_decomposition} it is clear that any movement of base learner predictions in opposing directions that leaves the joint loss unchanged must increase the mean individual loss accordingly. The diversity term therefore does not offer a ``free lunch'' to a jointly trained model on the training set: any apparent inflation of diversity under fixed joint prediction is paid for by a corresponding degradation of individual member performance.

The same decomposition admits a different reading. Equation~\eqref{eq:diversity_decomposition} can equally be rearranged as ${\mathrm{ERR}}_\text{ind} = \mathrm{ERR}_\text{joint} + \mathrm{DIV}$. Under this reading, it is the per-member objective that imposes the additional constraint, penalising diversity at fixed joint loss. Per-member training therefore systematically discourages individual members from making more extreme predictions, even in cases where their average would better match the target. Whether this constraint is desirable depends on the inductive bias of the model and training procedure.

Our experiments are consistent with this framing. Models trained with logit averaging produce per-trial predictions that are more extreme than would be optimal under per-trial training, mirroring the prediction-spreading behaviour reported by \citet{jeffares2023jointtrainingdeepensembles} for joint ensembles. Yet logit averaging \emph{succeeds} in our setting. Our results discussed in Section~\ref{sec:discussion} show that the diversity expressed at training time generalises to held-out data.

The success of logit averaging in our setting demonstrates that diversity learned under a joint objective can generalise. We suggest instead that the failure of joint ensemble training observed by \citet{jeffares2023jointtrainingdeepensembles} is more plausibly a consequence of the specific inductive biases of deep ensembles or of the loss landscape of the problem settings studied rather than an inherent issue with the joint training objective. Understanding why joint training succeeds in our multi-trial setting but fails in the ensemble setting is a useful direction for future work.

\subsection{Limitations}
\label{sec:discussion:limitations}

The experimental findings reported in this paper come with several limitations. First, our experiments are restricted to MEG recordings; we have not evaluated whether our findings transfer to EEG. Second, MegNIST is not yet publicly available, which limits the reproducibility of one of our seven experimental settings. We will release MegNIST in future work. Third, our analysis is limited to classifiers trained with cross-entropy loss. Whether our findings generalise to other loss functions, including contrastive losses such as SigLIP~\citet{zhai2023sigmoidlosslanguageimage}, is an open question for future work.

\section{Conclusion}
\label{sec:conclusion}
We have shown that aggregation choice matters substantially in non-invasive neural decoding, and that a simple change to the training objective, training-time logit averaging, is the most effective aggregation strategy across the settings we evaluate. On the phoneme classification task of the 2025 PNPL competition, it sets a new state of the art, outperforming the competition winner by 17.4 absolute percentage points. We further connect logit averaging to recent theoretical analyses of joint ensemble training, which has been shown to fail in the deep learning literature, and argue that our results call into question recent explanations of why this failure occurs. Understanding why joint training succeeds in our multi-trial setting but fails for ensembles remains a useful direction for future work. Together, these findings establish multi-trial aggregation as a key architectural choice in non-invasive neural decoding and identify training-time logit averaging as a simple, effective default.

\bibliographystyle{plainnat}
\bibliography{references}

\appendix


\section{Anonymized code}
\label{app:code}

The anonymised analysis code accompanying this submission is available at:
\begin{center}
\url{https://anonymous.4open.science/r/multisample-meg-B0D8/}
\end{center}

The repository contains the code used to run the experiments reported in this paper, including implementations of the aggregation paradigms, and training and evaluation scripts. Author-identifying metadata has been removed for double-blind review. A non-anonymised version will be made available upon publication. Each experiments can be reproduced on a single Nvidia H100 GPU in under 12 hours. 

\section{Datasets overview}
\label{app:datasets-table}

This appendix summarises the datasets and tasks evaluated in the paper. Each row in Table~\ref{tab:datasets} corresponds to one (dataset, task) combination, with group sizes indicating the number of trials aggregated per prediction.

\begin{table}[t]
\centering
\caption{Datasets and tasks evaluated in this paper. $K$ denotes the number of classes; $C \times T$ the number of sensor channels and time samples per trial; $g$ the group size, i.e.\ the number of trials aggregated per prediction. All recordings are downsampled to 250\,Hz.}
\label{tab:datasets}
\small
\begin{tabular}{llcccll}
\toprule
\textbf{Task} & \textbf{Dataset} & $K$ & $C \times T$ & $g$ & \textbf{Stimulus} \\
\midrule
Phoneme Classification     & LibriBrain & 39 & $306 \times 125$ & 10, 100 & Audiobook \\
Phoneme Voicing            & LibriBrain & 2  & $306 \times 125$ & 10      & Audiobook \\
Word Classification (top-10) & LibriBrain & 10 & $306 \times 250$ & 10      & Audiobook \\
Phoneme Classification     & Armeni     & 39 & $269 \times 125$ & 10      & Audiobook \\
Phoneme Voicing            & Armeni     & 2  & $269 \times 125$ & 10      & Audiobook \\
Digit Classification       & MegNIST    & 10 & $306 \times 250$ & 10      & Imagined Speech \\
\bottomrule
\end{tabular}
\end{table}

\section{Learning rate tuning}
\label{app:lr_tuning}

For each combination of dataset, paradigm, and group size, we tune the learning rate in two stages. We first evaluate three learning rates ($7 \times 10^{-5}$, $1 \times 10^{-4}$, $3 \times 10^{-4}$) with a single seed. Where the upper or lower edge of this initial grid performs noticeably better, we extend the grid in the relevant direction: $5 \times 10^{-4}$ and $1 \times 10^{-3}$ above the upper edge, and $4 \times 10^{-5}$ below the lower edge. Post-hoc averaging is trained once per (dataset, learning rate) and evaluated at all group sizes, since its training does not depend on the group size. The learning rate yielding the highest validation balanced accuracy is then fixed for the final evaluation.

\section{Architectural details for attention-based aggregation methods}
\label{app:attention_methods}

This appendix describes the three attention-based aggregation methods introduced in Section~\ref{sec:methods:proposed} in full architectural detail.

\paragraph{Shared framework.}
All three methods aggregate at the output of the backbone's first convolutional layer. Let $f_\theta^\text{front}$ denote this first layer, producing per-trial features $h_i = f_\theta^\text{front}(x_i) \in \mathbb{R}^{D \times T'}$, where $D$ is the feature dimension and $T'$ the time dimension after the first convolution. A pooling module aggregates the feature maps $\{h_1, \ldots, h_g\}$ into a single feature map $\bar{h} \in \mathbb{R}^{D \times T'}$, which is then passed through the remainder of the backbone $f_\theta^\text{rest}$ to produce a single set of logits $\hat{\ell} = f_\theta^\text{rest}(\bar{h})$. The methods differ only in the choice of pooling module.

\paragraph{Learned weighted averaging.}
Each per-trial feature map is summarised by averaging over the time dimension to produce a token $\tilde{h}_i \in \mathbb{R}^D$. A small MLP $\phi$ maps each token to a scalar score $s_i = \phi(\tilde{h}_i)$, and a softmax over trials yields normalised weights $w_i = \exp(s_i) / \sum_j \exp(s_j)$. The pooled feature map is the weighted sum of the original per-trial feature maps:
\begin{equation}
    \bar{h} = \sum_{i=1}^{g} w_i \, h_i.
\end{equation}

\paragraph{Cross-trial attention.}
Each per-trial feature map is again summarised into a token $\tilde{h}_i \in \mathbb{R}^D$ by averaging over time. The tokens are projected to an embedding space, normalised, and passed through multi-head self-attention so that each trial's representation is updated by attending to all other trials. The refined tokens are mapped to scalar scores via a linear projection, and a softmax over trials produces weights $w_i$ that are used to combine the original feature maps as in learned weighted averaging. This method differs from learned weighted averaging in that the per-trial scores depend on all other trials in the group, rather than on each trial in isolation.

\paragraph{Learned query attention.}
A small fixed number of trainable query vectors attend over the per-trial tokens $\{\tilde{h}_1, \ldots, \tilde{h}_g\}$ via multi-head attention. The attention output is projected back to dimension $D$ and broadcast across time, then combined with a naive elementwise mean of the per-trial feature maps via a learned gating mechanism:
\begin{equation}
    \bar{h} = \frac{1}{g} \sum_{i=1}^{g} h_i + \alpha \cdot \mathrm{Attn}(\{\tilde{h}_1, \ldots, \tilde{h}_g\}),
\end{equation}
where $\alpha = \sigma(\beta) \cdot w_t$ with $\beta$ a learnable scalar initialised at $-3$ (so $\sigma(\beta) \approx 0.05$ at the start of training) and $w_t$ a warmup weight that ramps linearly from $0$ to $1$ over the first ten epochs. This residual formulation allows the attention module to refine the simpler mean-pooled representation rather than replace it.

\section{Results for attention-based aggregation methods}
\label{app:attention_results}

This appendix reports the test balanced accuracy of the three attention-based aggregation methods introduced in Section~\ref{sec:methods:proposed} on the same conditions evaluated in Table~\ref{tab:main_results}. Architectural details for these methods are given in Appendix~\ref{app:attention_methods}. None of the three methods consistently outperforms training-time logit averaging across the conditions evaluated; we discuss possible reasons in Section~\ref{sec:discussion:attention}.

\begin{table}[h]
\centering
\caption{Test balanced accuracy of the three attention-based aggregation methods across all seven (dataset, task, group size) conditions. LWA: learned weighted averaging. CTA: cross-trial attention. LQA: learned query attention. Values are mean $\pm$ standard deviation across 5 random seeds. Significance markers indicate paired $t$-tests against input averaging: $^{\ast}$ for $p < 0.05$, $^{\ast\ast}$ for $p < 0.01$.}
\label{tab:attention_results}
\small
\setlength{\tabcolsep}{6pt}
\begin{tabular}{llccc}
\toprule
\textbf{Dataset} & \textbf{Task} & LWA & CTA & LQA \\
\midrule
\multirow{4}{*}{LibriBrain}
& Phoneme ($g=10$) & $.170 \pm .014$ & $.168 \pm .005$ & $.184 \pm .009$ \\
& Phoneme ($g=100$) & $.558 \pm .045$ & $.595 \pm .033$ & $.561 \pm .014$ \\
& Voicing ($g=10$) & $.792 \pm .006$ & $.792 \pm .013$ & $.778 \pm .016$ \\
& Word top-10 ($g=10$) & $.506 \pm .046$ & $.514 \pm .053$ & $.515 \pm .060$ \\
\midrule
MegNIST & Digit ($g=10$) & $.482 \pm .053^{\ast}$ & $.462 \pm .044$ & $.436 \pm .030$ \\
\midrule
\multirow{2}{*}{Armeni}
& Phoneme ($g=10$) & $.216 \pm .008$ & $.213 \pm .003$ & $.211 \pm .007$ \\
& Voicing ($g=10$) & $.785 \pm .002$ & $.789 \pm .004$ & $.788 \pm .002$ \\
\bottomrule
\end{tabular}
\end{table}

\section{Per-class analysis on LibriBrain phoneme classification}
\label{app:perclass}

This appendix examines per-class performance on the LibriBrain phoneme classification task to better characterise the failure mode of post-hoc averaging discussed in Section~\ref{sec:experiments:main} and complements the calibration analysis in Appendix~\ref{app:calibration}.

\paragraph{Setup.}
We split the 39 phoneme classes into three tertiles by training-set frequency: low-frequency (the 13 least common classes, with 605--15{,}086 training samples each), mid-frequency (ranks 14--26, 18{,}013--49{,}851 samples), and high-frequency (the 13 most common, 51{,}729--158{,}318 samples). For each tertile we report macro recall, macro $F_1$, and the mean predicted probability on the true class $\bar p_{\text{true}} = \mathbb{E}_{(x, y) \sim D}\, p_\theta(y \mid x)$, averaged across 5 seeds. 

\begin{table}[h]
\centering
\small
\setlength{\tabcolsep}{4pt}
\begin{tabular}{lll ccc ccc ccc}
\toprule
& & & \multicolumn{3}{c}{Macro recall} & \multicolumn{3}{c}{Macro $F_1$} & \multicolumn{3}{c}{$\bar p_{\text{true}}$} \\
\cmidrule(lr){4-6} \cmidrule(lr){7-9} \cmidrule(lr){10-12}
Paradigm & $g$ & & low & mid & high & low & mid & high & low & mid & high \\
\midrule
Input averaging   & 10  && 0.145 & 0.165 & 0.214 & 0.085 & 0.159 & 0.212 & 0.098 & 0.100 & 0.110 \\
Logit averaging   & 10  && \textbf{0.201} & \textbf{0.239} & \textbf{0.277} & \textbf{0.141} & \textbf{0.225} & \textbf{0.287} & \textbf{0.131} & \textbf{0.151} & \textbf{0.156} \\
Post-hoc averaging & 10  && 0.000 & 0.008 & 0.207 & 0.000 & 0.014 & 0.130 & 0.023 & 0.034 & 0.047 \\
\midrule
Input averaging   & 100 && 0.667$^\dagger$ & 0.471 & 0.636 & 0.113$^\dagger$ & 0.410 & 0.634 & 0.666$^\dagger$ & 0.455 & 0.608 \\
Logit averaging   & 100 && 0.600$^\dagger$ & \textbf{0.601} & \textbf{0.690} & \textbf{0.126}$^\dagger$ & \textbf{0.531} & \textbf{0.693} & 0.536$^\dagger$ & \textbf{0.577} & \textbf{0.657} \\
Post-hoc averaging & 100 && 0.000 & 0.000 & 0.251 & 0.000 & 0.000 & 0.185 & 0.029 & 0.034 & 0.047 \\
\bottomrule
\end{tabular}
\caption{Per-class performance on phoneme classification, broken out by training-frequency tertile. The best value in each (metric, $g$, tertile) cell is shown in \textbf{bold}. $\bar p_{\text{true}}$ denotes the mean predicted probability assigned to the true class. $^\dagger$The low-frequency tertile at $g{=}100$ contains only 3 of the 13 classes assigned to the bucket; the remaining 10 have insufficient test samples to form complete groups and are excluded. Values are means over 5 seeds.}
\label{tab:perclass-tertile}
\end{table}

\paragraph{Logit averaging improves uniformly across class frequencies.}
At $g{=}10$, where every tertile is evaluated on the same class set, logit averaging improves macro recall over input averaging by $+0.056$, $+0.074$, and $+0.063$ on the low-, mid-, and high-frequency tertiles respectively. The improvement is broadly flat: logit averaging does not selectively rescue rare classes but provides a consistent gain across the frequency spectrum, with a mild peak in the middle tertile. The $\bar p_{\text{true}}$ values show the same shape ($+0.033$, $+0.051$, $+0.046$). At $g{=}100$, the gain over input averaging is concentrated in the mid-frequency tertile ($\Delta\bar p_{\text{true}} = +0.122$): with enough samples per group, input averaging recovers most of the noise-reduction benefit on the easiest and hardest classes, leaving intermediate-difficulty classes as the regime where logit averaging retains a clear advantage. 

\paragraph{Post-hoc averaging collapses on minority classes.}
Post-hoc averaging exhibits a distinctive failure mode on phoneme classification. At $g{=}10$, its macro recall on the low- and mid-frequency tertiles is $0.000$ and $0.008$ respectively; at $g{=}100$, both are $0.000$. The model produces non-zero recall \emph{only} on the high-frequency tertile, and even there inspection of per-class predictions reveals that post-hoc effectively predicts only the four most common phonemes ($\{2, 22, 28, 30\}$) for every test group.

This collapse follows directly from the averaging rule. Post-hoc averaging combines bounded per-trial probabilities $p_n \in [0, 1]^C$, so a class with low marginal probability across all $g$ trials cannot accumulate enough probability mass to overtake a more frequent alternative, even when it is the correct answer. 

\section{Calibration analysis}
\label{app:calibration}

This appendix reports a more detailed analysis of model calibration referenced in Section~\ref{sec:discussion:mechanism}. Calibration is assessed for models trained with training-time logit averaging and with post-hoc averaging, evaluated both on single ungrouped trials (Table~\ref{tab:single_trial_calibration}) and at the group level used during evaluation (Table~\ref{tab:group_calibration}). All calibration metrics use expected calibration error (ECE), where lower values indicate better agreement between predicted probabilities and empirical correctness rates.

\subsection{Single-trial calibration}

When applied to ungrouped trials, models trained with logit averaging exhibit substantially worse calibration than models trained with per-trial cross-entropy (Table~\ref{tab:single_trial_calibration}). ECE values for logit averaging range from $0.058$ on phoneme voicing to $0.899$ on phoneme classification, while post-hoc models maintain ECE below $0.16$ on every dataset. This is consistent with the loss-landscape argument in Section~\ref{sec:discussion:mechanism}: training-time logit averaging does not optimise for honest per-trial probabilities, and the resulting models produce single-trial logits whose magnitudes substantially exceed what per-trial calibration would warrant.

Per-trial balanced accuracy is not uniformly worse for logit averaging, however. On phoneme classification and phoneme voicing, single-trial balanced accuracy is comparable or higher for logit averaging despite the calibration gap. On MegNIST and word top-10, post-hoc retains a per-trial advantage. The per-trial accuracy results are not directly tied to the calibration story --- they reflect what the per-trial features have learned, not how confident the model is about them.

\begin{table}[h]
\centering
\caption{Single-trial evaluation of trained models, applied to ungrouped trials. Values are mean $\pm$ standard deviation across 5 seeds. Lower ECE indicates better calibration of single-trial probabilities.}
\label{tab:single_trial_calibration}
\small
\begin{tabular}{lcccc}
\toprule
& \multicolumn{2}{c}{\textbf{Balanced accuracy}} & \multicolumn{2}{c}{\textbf{ECE}} \\
\cmidrule(lr){2-3} \cmidrule(lr){4-5}
\textbf{Dataset} & Logit avg.\ & Post-hoc & Logit avg.\ & Post-hoc \\
\midrule
MegNIST          & $0.196 \pm 0.009$ & $0.245 \pm 0.002$ & $0.715 \pm 0.009$ & $0.156 \pm 0.034$ \\
Phoneme          & $0.060 \pm 0.002$ & $0.052 \pm 0.003$ & $0.899 \pm 0.011$ & $0.021 \pm 0.004$ \\
Phoneme voicing  & $0.600 \pm 0.002$ & $0.553 \pm 0.003$ & $0.276 \pm 0.006$ & $0.071 \pm 0.015$ \\
Word top-10      & $0.272 \pm 0.019$ & $0.308 \pm 0.006$ & $0.626 \pm 0.045$ & $0.047 \pm 0.008$ \\
\bottomrule
\end{tabular}
\end{table}

\subsection{Group-level calibration}

The picture changes when models are evaluated at the group size used during training and at test time ($g=10$ in this analysis). At the group level, logit averaging is comparably or better calibrated than post-hoc on three of four tasks (Table~\ref{tab:group_calibration}). On phoneme classification, group-level ECE for logit averaging ($0.036$) is roughly four times lower than for post-hoc ($0.124$). On phoneme voicing and word top-10, the gap is smaller but still favours logit averaging. Only on MegNIST does logit averaging remain less calibrated than post-hoc at the group level.

\begin{table}[h]
\centering
\caption{Group-level evaluation at $g=10$ for the same models. Balanced accuracy reproduces values from Table~\ref{tab:main_results}; ECE is reported for completeness. Values are mean $\pm$ standard deviation across 5 seeds.}
\label{tab:group_calibration}
\small
\begin{tabular}{lcccc}
\toprule
& \multicolumn{2}{c}{\textbf{Balanced accuracy}} & \multicolumn{2}{c}{\textbf{ECE}} \\
\cmidrule(lr){2-3} \cmidrule(lr){4-5}
\textbf{Dataset} & Logit avg.\ & Post-hoc & Logit avg.\ & Post-hoc \\
\midrule
MegNIST          & $0.418 \pm 0.008$ & $0.520 \pm 0.047$ & $0.364 \pm 0.010$ & $0.285 \pm 0.036$ \\
Phoneme          & $0.241 \pm 0.011$ & $0.066 \pm 0.005$ & $0.036 \pm 0.005$ & $0.124 \pm 0.010$ \\
Phoneme voicing  & $0.794 \pm 0.011$ & $0.510 \pm 0.005$ & $0.058 \pm 0.011$ & $0.185 \pm 0.021$ \\
Word top-10      & $0.564 \pm 0.019$ & $0.403 \pm 0.025$ & $0.162 \pm 0.078$ & $0.272 \pm 0.038$ \\
\bottomrule
\end{tabular}
\end{table}

\subsection{Confidence-ranked accuracy}

A complementary view of group-level calibration is given by ranking predictions by confidence and measuring per-decile accuracy (Table~\ref{tab:decile_accuracy}). For logit averaging, the most-confident decile is substantially more accurate than the least-confident decile across all tasks: on phoneme classification, balanced accuracy in the top decile is $0.426$ versus $0.128$ in the bottom decile; on phoneme voicing, $0.972$ versus $0.522$. Post-hoc shows weaker confidence ordering on the speech tasks. Most strikingly, on phoneme voicing, post-hoc balanced accuracy is essentially identical ($\approx 0.500$) across the top nine deciles --- the model's confidence rankings provide no useful information about which predictions are likely to be correct. Logit averaging produces confidence rankings that translate into clear accuracy gradients, even on tasks where its single-trial calibration is poor.

\begin{table}[h]
\centering
\caption{Per-decile balanced accuracy for group-level predictions ($g=10$), ranked by group-level confidence (decile 1: most confident; decile 10: least confident). Values are means across 5 seeds.}
\label{tab:decile_accuracy}
\small
\begin{tabular}{llcccccccccc}
\toprule
\textbf{Dataset} & \textbf{Method} & 1 & 2 & 3 & 4 & 5 & 6 & 7 & 8 & 9 & 10 \\
\midrule
\multirow{2}{*}{MegNIST} & Logit avg.\ & 0.86 & 0.51 & 0.43 & 0.44 & 0.38 & 0.34 & 0.28 & 0.23 & 0.32 & 0.24 \\
                         & Post-hoc    & 0.96 & 0.81 & 0.76 & 0.63 & 0.51 & 0.40 & 0.30 & 0.34 & 0.30 & 0.23 \\
\addlinespace
\multirow{2}{*}{Phoneme} & Logit avg.\ & 0.43 & 0.36 & 0.33 & 0.26 & 0.18 & 0.22 & 0.19 & 0.16 & 0.15 & 0.13 \\
                         & Post-hoc    & 0.14 & 0.12 & 0.09 & 0.08 & 0.08 & 0.07 & 0.05 & 0.07 & 0.07 & 0.06 \\
\addlinespace
\multirow{2}{*}{Phoneme voicing} & Logit avg.\ & 0.97 & 0.97 & 0.88 & 0.92 & 0.87 & 0.81 & 0.77 & 0.73 & 0.67 & 0.52 \\
                                 & Post-hoc    & 0.70 & 0.50 & 0.50 & 0.50 & 0.50 & 0.50 & 0.50 & 0.50 & 0.50 & 0.52 \\
\addlinespace
\multirow{2}{*}{Word top-10} & Logit avg.\ & 0.90 & 0.88 & 0.78 & 0.73 & 0.44 & 0.55 & 0.51 & 0.38 & 0.28 & 0.34 \\
                             & Post-hoc    & 0.93 & 0.73 & 0.55 & 0.34 & 0.46 & 0.42 & 0.36 & 0.37 & 0.38 & 0.23 \\
\bottomrule
\end{tabular}
\end{table}

\subsection{Interpretation}

The single-trial and group-level calibration results together suggest that the per-trial logits produced by logit averaging are not independently calibrated, but combine constructively under averaging in a way that single-trial training does not produce. A model trained with per-trial cross-entropy can produce per-trial probabilities that are individually well-calibrated and yet, when averaged, fail to provide useful discrimination between confident and uncertain predictions at the group level.

\newpage

\end{document}